# AUTOMATED WORDS STABILITY AND LANGUAGES PHYLOGENY

**Filippo Petroni**[(1)] **and Maurizio Serva**[(2)]

[(1)]*Dipartimento di Matematica, Facoltà di Economia,*
*Università la Sapienza, via del Castro Laurenziano,9*
*00161 Roma*
[(2)]*Dipartimento di Matematica,*
*Università dell'Aquila, I-67010 L'Aquila, Italy*

**ABSTRACT**

The idea of measuring distance between languages seems to have its roots in the work of the French explorer Dumont D'Urville (D'Urville 1832). He collected comparative words lists of various languages during his voyages aboard the Astrolabe from 1826 to1829 and, in his work about the geographical division of the Pacific, he proposed a method to measure the degree of relation among languages. The method used by modern glottochronology, developed by Morris Swadesh in the 1950s (Swadesh 1952), measures distances from the percentage of shared cognates, which are words with a common historical origin. Recently, we proposed a new automated method which uses normalized Levenshtein distance among words with the same meaning and averages on the words contained in a list.

Another classical problem in glottochronology is the study of the stability of words corresponding to different meanings. Words, in fact, evolve because of lexical changes, borrowings and replacement at a rate which is not the same for all of them. The speed of lexical evolution is different for different meanings and it is probably related to the frequency of use of the associated words (Pagel *et al*. 2007). This problem is tackled here by an automated methodology only based on normalized Levenshtein distance.

**INTRODUCTION**

Glottochronology tries to estimate the time at which languages diverged with the implicit assumption that vocabularies change at a constant average rate. The concept seems to have its roots in the work of the French explorer Dumont D'Urville. He collected comparative words lists of various languages during his voyages aboard the Astrolabe from 1826 to 1829 and, in his work about the geographical division of the Pacific (D'Urville 1832) he introduced the concept of lexical cognates and proposed a method to measure the degree of relation among languages. He used a core vocabulary of 115 basic terms which, impressively, contains all but three the terms of the Swadesh 100-item list. Then, he assigned a distance from 0 to 1 to any pair of words with same meaning and finally he was able to resolve the relationship for any pair of languages. His conclusion is famous: La langue est partout la meme.

The method used by modern glottochronology, was developed by Morris Swadesh (Swadesh 1952) in the 1950s. The idea is to consider the percentage of shared cognates in order to compute the distance between pairs of languages. These lexical distances are assumed to be, on average, logarithmically proportional to divergence times. In fact, changes in vocabulary accumulate year after year and two languages initially similar become more and more different. A recent example of the use of Swadesh lists and cognates to construct language trees are the studies of Gray and Atkinson (Gray and Atkinson 2003) and Gray and Jordan (Gray and Jordan (2000)).

We recently proposed an automated method which uses Levenshtein distance among words in a list (Serva and Petroni 2008, Petroni and Serva 2008). To be precise, we defined the distance of two languages by considering a normalized Levenshtein distance among words with the same meaning and we averaged on all the words contained in a list[1]. The normalization, which takes into account word length, plays a crucial role, and no sensible results would have been found without. We applied our method to the Indo-European and the Austronesian groups considering, in both cases, fifty different languages (Serva and Petroni 2008, Petroni and Serva 2008).

Almost at the same time, the above described automated method was used and developed by another large group of scholars (Bakker *et al.* 2008, Holman *et al.* 2008). In their work, they used lists of 40 words while we used lists of 200. Their choice was taken according to a careful study of the stability of different words (Wichmann 2009).

Another classical problem in glottochronology is the study of the stability of words corresponding to different meanings. Words in fact, evolve because of lexical changes, borrowings and replacement at a rate which is not the same for all of them. The speed of lexical evolution, is different for different meanings and it is probably related to the frequency of use of the associated words (Pagel *et al*. 2007) . The study of words stability has an interest in itself since it may give strong information on the activities which are at the core of the behavior of a social or ethnic group but it is also necessary for a proper choice of the imput lists for language comparisons.

The idea of inferring the stability of an item from its similarity in related languages goes back a long way in the lexicostatistical literature (Thomas 1960, Kroeber 1963, Oswalt 1971). In this paper we tackle this problem with an automated methodology based on normalized Levenshtein distance. To reach the goal, it is necessary to obtain a measure of the typical distance of all pairs of words corresponding to a given meaning in a language family.
The distance between words is computed as in (Serva and Petroni 2008, Petroni and Serva 2008) avoiding the use of cognates. For any meaning, and any language family, we are able to find a number which measure its stability (or rate of evolution) in a completely objective and reproducible manner.

In the next section we define the lexical distance between words. Section 3 is the core of the paper, there we define the automated stability of the meanings and we study the distribution and ranking of stability for Indo-European family and for Austronesian one. In section 4 we compare the stability ranking of items. Conclusions and outlook are in section 5.

**LEXICAL DISTANCE**

Our definition of lexical distance between two words is a variant of the Levenshtein distance which is simply the minimum number of insertions, deletions, or substitutions of a single character needed to transform one word into the other. Our definition is taken as the Levenshtein distance divided by the number of characters of the longer of the two compared words.

More precisely, given two words $α_i$ and $β_i$ corresponding to the same item $i$ in two languages $α$ and $β$, their distance $D(α_i,β_i)$ is given by

$$D(α_i, β_i) = \frac{D_l(α_i, β_i)}{L(α_i, β_i)}$$

---

[1] The database, modified by the Authors, is available at the following web address: http://univaq.it/~serva/languages/languages.html. Readers are welcome to modify, correct and add words to the database.

where $D_l(α_i,β_i)$ is the Levenshtein distance between the two words and $L(α_i,β_i)$ is the number of characters of the longer of the two words $α_i$ and $β_i$. Therefore, the distance can take any value between 0 and 1. Obviously $D(α_i,α_i)= 0$ since both the item and the language coincide.

The normalization is an important novelty and it plays a crucial role; no sensible results can been found without. The reason why we normalize can be understood from the following example. Consider the case in which a single substitution transforms one word into the other with the same length. If they are short, let's say 2 characters, they are very different. On the contrary, if they are long, let's say 8 characters, it is reasonable to say that they are very similar. Without normalization, their distance would be the same and equal to 1, regardless of their length. Instead, introducing the normalization factor, in the first case the distance is 1/2, whereas in the second, it is much smaller and equal to 1/8.

In (Serva and Petroni 2008, Petroni and Serva 2008) we used distance between pairs of words, as defined above, to construct the lexical distances of languages. For any pair of languages, the first step was to compute the distance between words corresponding to the same meaning. The lexical distance between each pair of languages was defined as the average of the distance between all words in the Swadesh list. As a result we obtained a number between 0 and 1 which is the lexical distance between two languages. Then, we performed a logarithmic transformation of lexical distances into separation times with an analogous of the adjusted fundamental formula of glottochronology (Starostin 1999). Finally, the phylogenetic trees for the Austronesian and Indo-European families could be straightforwardly constructed.

Criticism has been made to our proposal (Nichols and Warnow 2008) on the basis that our reconstructed trees present some incongruence as for example the early separation of Armenian which is not grouped together with Greek (which in our Indo-European tree separate just after Armenian). Nevertheless, the structure of the top of the tree is debated and no universally accepted conclusion exists.

**STABILITY OF MEANINGS**

We take now decisions concerning stability of meanings. Our aim is to obtain an automated procedure, which avoids, also at this level, the use of cognates. For this purpose, it is necessary to obtain a measure of the typical distance of all pairs of words corresponding to a given meaning in a language family.

Assume that the number of languages in the considered family is $N$ and the list of words for any language contains $M=200$ items. Any language in the group is labeled with a Greek letter (say $α$) and any word of that language by $α_i$ with $1≤i≤ M$. Then, two words $α_i$ and $β_i$ in the languages $α$ and $β$ have the same meaning.

Therefore, we define the stability as:

$$S(i) = 1 - \frac{1}{N(N-1)} \sum_{α \neq β} D(α_i, β_i)$$

where the sum goes on all possible $N(N-1)$ language pairs $α, β$ in the family.

With this definition, $S(i)$ is inversely proportional to the average of the distances $D(α_i,β_i)$ and it takes a value between 0 and 1. The averaged distance is smaller for those words corresponding to meanings with a lower rate of lexical evolution since they tend to remain more similar in two languages. Therefore, to a larger $S(i)$ corresponds a greater stability.

We computed the $S(i)$ for 200 meanings averaging over 50 languages of the Indo-European family and the same for of the Austronesian one.

To have a first qualitative understanding we plot the two associated histograms shown in Fig 1. We can see that, in both cases, there is a fat tail on the right of the histograms indicating that there are some meaning with a quite large stability. This tail is at very variance with a standard Gaussian behavior.

We remark that similar plots were computed in (Pagel et al. 2007) were the rates of lexical evolution are obtained by the standard glottochronology approach.

To understand better the behavior of the stability distribution, we plot $S(i)$, in a decreasing rank, for the 200 meaning taken from the Swadesh. In Fig. 2 we report the data concerning Indo-European family and Austronesian families. For both families, the stability drops rapidly at the beginning, then, between the 50th position and the 180th it decreases slowly and almost linearly with rank, finally at the end stability drops again. In both figures we fit by a straight line the central part of the data between position 51 and position 180, in order to highlight the initial and final deviation from the linear behavior. One can easily conclude that both families have 50 meanings with a particularly high information content and 20 meanings with particularly low one.

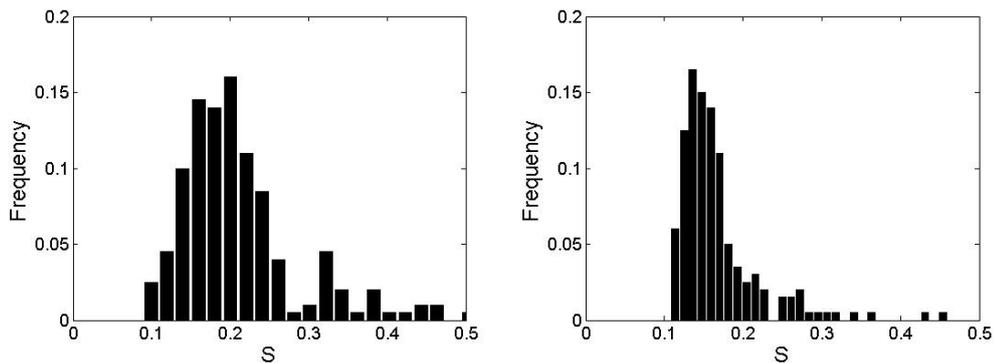

*Figure 1: Stability histogram of meanings for Austronesian (left) and Indo-European (right) languages. The fat tail on the right of the histograms indicates that some items have a very large stability. The qualitative behavior of the two families is the same.*

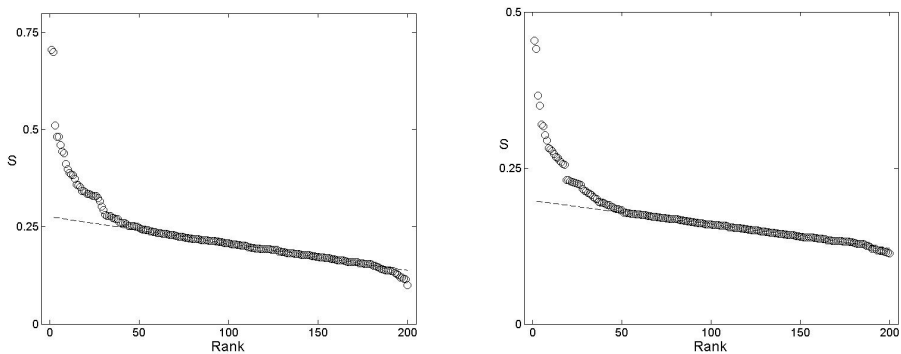

*Figure 2: Stability in a decreasing rank for the 200 meanings of the Austronesian (left) and Indo-European (right) languages. At the beginning stability has large values but drops rapidly, then, between the $50^{th}$ and the $180^{th}$ position it decreases linearly, finally it drops again. The straight line between position 51 and position 180 underlines the initial and final deviation from the linear behavior. Also in this case the qualitative behavior of the two groups is the same.*

**COMPARISON**

We study the stability correlations between same items in the two language families and we also compare the stability ranking of items. We found that the correlation coefficient between the stability index computed for the two groups is roughly 0.21. This number is positive and it evidences a certain correlation between ranking in the two families. Nevertheless, its low value suggest that the stability of items depends strongly on the studied family. Only looking at the overall correlation we are not able to understand its origin, since it could be a consequence of the strong correlation of few items or a week correlation of many of them. In other words, it could be that the most stable terms in the two list show a large coincidence, while the other a lower or vanishing one. Or it also could be possible that a small coincidence can be found both for very stable and low stable items.

To better understand this point we considered the first *n* items in the ranking list for both families, and we computed the number *m(n)* of common items in the two lists. To underline the non casual behavior, *m(n)* has to be compared with $n^2/N$ which is the average number of common items if one randomly chooses *n* items from any of the two lists. Then, it is natural to define *p(n)* as *m(n)* divided by $n^2/N$. If there is no relation between stability in the two families *p(n)* must be close to 1 for every *n*. The behavior of *p(n)* as a function of *n* can be seen in figure 3 which shows that indeed there is a non trivial overlapping of the two lists of most stable *n* items since *p(n)* is always larger than 1. This fact confirms the correlation between the two rankings, but also shows that this effect is strong only for small *n* (*n* less than 50). For larger *n* the overlapping is much closer to 1 and random coincidences prevail. This means that the most stable terms in the two list are those to show a larger coincidence.

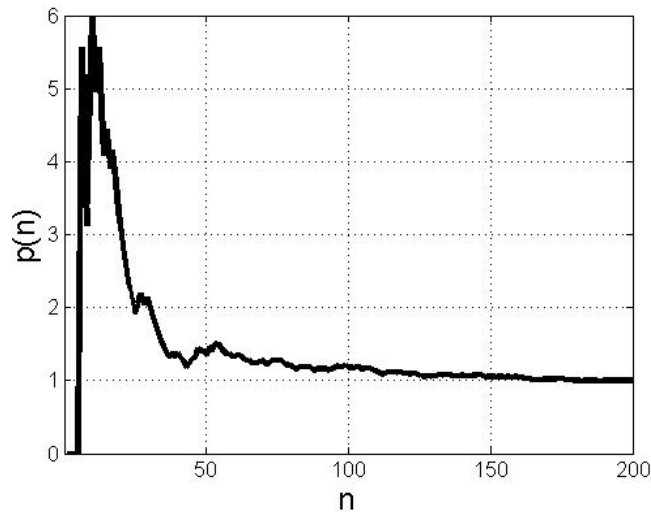

*Figure 3: In this figure it is shown the number of common items in the two list of most stable n items obtained for the Austronesian and Indo-European families. The number is normalized by the random coincidence $n^2/200$.*

To give an example of the lists found with our approach we show here a table of the 20 most stable items for the Indo-European and Austronesian languages groups. Together with any of the items we report its stability record within the family.

| INDO-EUROPEAN | S(i) | AUSTRONESIAN | S(i) |
|---|---|---|---|
| YOU | 0.45395 | EYE | 0.70646 |
| THREE | 0.44102 | FIVE | 0.70089 |
| MOTHER | 0.36627 | FATHER | 0.51095 |
| NOT | 0.35033 | DIE | 0.48157 |
| NEW | 0.31961 | STONE | 0.48157 |
| NOSE | 0.3169 | THREE | 0.46087 |
| FOUR | 0.30226 | TWO | 0.44411 |
| NIGHT | 0.29403 | LOUSE | 0.43958 |
| TWO | 0.28214 | ROAD | 0.41217 |
| NAME | 0.27962 | FOUR | 0.39798 |
| TOOTH | 0.27677 | HAND | 0.38997 |
| STAR | 0.27269 | NAME | 0.38493 |
| SALT | 0.26792 | LIVER | 0.38375 |
| DAY | 0.26695 | PUSH | 0.37444 |

| GRASS | 0.26231 | MOTHER | 0.35821 |
| SEA | 0.25906 | WE | 0.35749 |
| DIE | 0.25602 | EAT | 0.3529 |
| SUN | 0.25535 | STICK | 0.34242 |
| ONE | 0.23093 | I | 0.34208 |
| FEATHER | 0.23055 | VOMIT | 0.33861 |

*Table 1: The table shows the 20 most stable words for the Indo-European and Austronesian language groups. Together with any of the items we report its stability record within the family.*

**CONCLUSIONS**

The novelty of the approach we have proposed is that everything can be made automatically. One has only to choose a group of languages for which the relative lists of words exist. Then, stability can be computed automatically by using simple objective arguments. We do not claim that our method produces better results of the standard glottochronology approach, but surely comparable. The advantage being only that it avoids subjectivity since all results can be replicated by other scholars assuming that the database is the same. Furthermore, it allows for rapid comparison of items of a very large number of languages. In fact, the only work is to prepare the lists, while all the remaining work is made by a computer program. In this way the difficult and lengthy task of cognates identification is avoided.

We applied here our method to the Indo-European and Austronesian families of languages considering 200 items lists of words according to the original choice of Swadesh. The output was a stability measure for all items computed separately for the two families. The histogram of stability shows identical qualitative behavior in the two cases with a fat tail corresponding to items with very high stability. The ranking plot also shows that the two families behave in the same way, with the higher stability items deviating from the linear interpolation because of their very large values. We are convinced that this phenomenology we observe, both for Indo-European and Austronesian languages, should be a universal characteristic of stability distributions, common to all families. On the contrary, it turns out that the most stable items are not the same even if there is a positive correlation between the stability computed for Indo-European and Austronesian groups. We do not know, at this stage, why items may be stable within a family and unstable in one other. We only can speculate, according to recent study (Pagel *et al.* 2007), that this is related to the different frequency of use of words in different cultural contexts.

**ACKNOWLEDGMENTS**

We warmly thank Soeren Wichmann for helpful discussion. We also thank Philippe Blanchard, Armando Neves, Luce Prignano and Dimitri Volchenkov for critical comments on many aspects of the paper. We are indebted with S.J. Greenhill, R. Blust and R.D.Gray, for the authorization to use their: The Austronesian Basic Vocabulary Database.